%% file: main.tex
\documentclass[11pt,letterpaper]{article}
\usepackage{naaclhlt2013}
\usepackage{times}
\usepackage{latexsym}
\usepackage{amsmath}
\usepackage{mathtools}
\usepackage{amsfonts}
\usepackage{multirow}
\usepackage{url}
\usepackage{graphicx}
\usepackage{paralist}
\usepackage[nooneline]{subfigure}

\newcommand{\eat}[1]{\ignorespaces} 

\title{Probabilistic Frame Induction\thanks{This is a postprint version of a paper to appear in the \textit{Proceedings of the 2013 Conference of the North American Chapter of the Association for Computational Linguistics: Human Language Technologies (NAACL/HLT 2013)}.}}

\author{Jackie Chi Kit Cheung\thanks{This research was undertaken during the author's internship at Microsoft Research.} \\
   Department of Computer Science\\
   University of Toronto\\
   Toronto, ON, M5S 3G4, Canada\\
  {\small \tt jcheung@cs.toronto.edu} \\ \And
   Hoifung Poon\\
   One Microsoft Way\\
   Microsoft Research\\
   Redmond, WA 98052, USA\\
  {\small \tt hoifung@microsoft.com} \\ \And
   Lucy Vanderwende\\
   One Microsoft Way\\
   Microsoft Research\\
   Redmond, WA 98052, USA\\
  {\small \tt lucyv@microsoft.com}
  }
\date{}

\begin{document}
\maketitle

\input{abstract}

\input{intro}

\input{related}

\input{model}

\input{experiment}

\input{experiment2}

\input{summary}

\section*{Acknowledgments}
We would like to thank Nate Chambers for answering questions about his system. We would also like to thank Chris Quirk for help with preprocessing the MUC corpus, and the other members of the NLP group at Microsoft Research for useful discussions.

\bibliographystyle{naaclhlt2013}
\bibliography{cckitpw}

\end{document}

%% file: abstract.tex
\begin{abstract}
In natural-language discourse, related events tend to appear near each other to describe a larger scenario. Such structures can be formalized by the notion of a \textit{frame} (a.k.a. template), which comprises a set of related events and prototypical participants and event transitions. Identifying frames is a prerequisite for information extraction and natural language generation, and is usually done manually. Methods for inducing frames have been proposed recently, but they typically use ad hoc procedures and are difficult to diagnose or extend. In this paper, we propose the first probabilistic approach to frame induction, which incorporates frames, events, participants as latent topics and learns those frame and event transitions that best explain the text. The number of frames is inferred by a novel application of a split-merge method from syntactic parsing. In end-to-end evaluations from text to induced frames and extracted facts, our method produced state-of-the-art results while substantially reducing engineering effort.


\end{abstract}

%% file: intro.tex
\section{Introduction}
Events with causal or temporal relations tend to occur near each other in text. For example, a bombing scenario in an article on terrorism might begin with a {\tt DETONATION} event, in which terrorists set off a bomb. Then, a {\tt DAMAGE} event might ensue to describe the resulting destruction and any casualties, followed by an {\tt INVESTIGATION} event covering subsequent police investigations. Afterwards, the bombing scenario may transition into a criminal-processing scenario, which begins with police catching the terrorists, and proceeds to a trial, sentencing, etc.
A common set of participants serves as the event arguments; e.g., the agent (or subject) of {\tt DETONATION} is often the same as the theme (or object) of {\tt INVESTIGATION} and corresponds to the {\tt PERPETRATOR}.

Such structures can be formally captured by the notion of a {\em frame} (a.k.a. template), which consists of a set of {\em events} with prototypical transitions, as well as a set of {\em slots} representing the common participants. 
Identifying frames is an explicit or implicit prerequisite for many NLP tasks. Information extraction, for example, stipulates the types of events and slots that are extracted for a frame or template.
Online applications such as dialogue systems and personal-assistant applications also model users' goals and subgoals using frame-like representations, and in natural-language generation, frames are often used to represent content to be expressed as well as to support surface realization.

Until recently, frames and related representations have been manually constructed, which has limited their applicability to a relatively small number of domains and a few slots within a domain. Furthermore, additional manual effort is needed after the frames are defined in order to extract frame components from text (e.g., in annotating examples and designing features to train a supervised learning model). This paradigm makes it hard to generalize across tasks and might suffer from annotator bias.

Recently, there has been increasing interest in automatically inducing frames from text. A notable example is \newcite{chambers-jurafsky-2011}, which first clusters related verbs to form frames, and then clusters the verbs' syntactic arguments to identify slots. While \newcite{chambers-jurafsky-2011} represents a major step forward in frame induction, it is also limited in several aspects. The clustering used ad hoc steps and customized similarity metrics, as well as an additional retrieval step from a large external text corpus for slot generation. This makes it hard to replicate their approach or adapt it to new domains. Lacking a coherent model, it is also difficult to incorporate additional linguistic insights and prior knowledge.

In this paper, we present \texttt{ProFinder} (PRObabilistic Frame INDucER), which is the first probabilistic approach for frame induction. 
\texttt{ProFinder} defines a joint distribution over the words in a document and their frame assignments by modeling frame and event transition, correlations among events and slots, and their surface realizations.
Given a set of documents, \texttt{ProFinder} outputs a set of induced frames with learned parameters, as well as the most probable frame assignments that can be used for event and entity extraction.
The numbers of events and slots are dynamically determined by a novel application of the split-merge approach from syntactic parsing \cite{petrov-etal-2006}.
In end-to-end evaluations from text to entity extraction using the standard MUC and TAC datasets, ProFinder achieved state-of-the-art results while significantly reducing engineering effort and requiring no external data.

%% file: related.tex
\section{Related Work}
In information extraction and other semantic processing tasks, the dominant paradigm requires two stages of manual effort.
First, the target representation is defined manually by domain experts. Then, manual effort is required to construct an extractor or annotate examples to train a machine-learning system.
Recently, there has been a burgeoning body of work in alleviating such manual effort.
For example, a popular approach to reduce annotation effort is bootstrapping from seed examples \cite{patwardhan-riloff-2007,huang-riloff-2012}. However, this still requires prespecified frames or templates, and selecting seed words is often a challenging task due to semantic drift \cite{curran-etal-2007}.
Open IE \cite{banko-etzioni-2008} reduces the manual effort to designing a few domain-independent relation patterns, which can then be applied to extract relational triples from text. While extremely scalable, this approach can only extract atomic factoids within a sentence, and the resulting triples are noisy, non-cannonicalized text fragments.

More relevant to our approach is the recent work in unsupervised semantic induction, such as unsupervised semantic parsing \cite{poon-domingos-2009}, unsupervised semantical role labeling \cite{swier-stevenson-2004} and induction \cite[e.g.]{lang-lapata-2011}, and slot induction from web search logs \cite{cheung-li-2012}.
As in \texttt{ProFinder}, they also model distributional contexts for slot or role induction. However, these approaches focus on semantics in independent sentences, and do not capture discourse-level dependencies.

The modeling component for frame and event transitions in \texttt{ProFinder} is similar to a sequential topic model \cite{gruber-etal-2007}, and is inspired by the successful applications of such topic models in summarization \cite[inter alia]{barzilay-lee-2004,daume-marcu-2006,haghighi-vanderwende-2009}.
There are, however, two main differences. First, \texttt{ProFinder} contains not a single sequential topic model, but two (for frames and events, respectively). In addition, it also models the interdependencies among events, slots, and surface text, which is analogous to the USP model \cite{poon-domingos-2009}. ProFinder can thus be viewed as a novel combination of state-of-the-art models in unsupervised semantics and discourse modeling.

In terms of aim and capability, \texttt{ProFinder} is most similar to \newcite{chambers-jurafsky-2011}, which culminated from a series of work for identifying correlated events and arguments in narrative \cite{chambers-jurafsky-2008,chambers-jurafsky-2009}.
By adopting a probabilistic approach, \texttt{ProFinder} has a sound theoretical underpinning, and is easy to modify or extend. For example, in Section 3, we show how \texttt{ProFinder} can easily be augmented with additional linguistically-motivated features. Likewise, \texttt{ProFinder} can easily be used as a semi-supervised system if some slot designations and labeled examples are available. 

The idea of representing and capturing stereotypical knowledge has a long history in artificial intelligence and psychology, and has assumed various names such as \textit{frames} \cite{minsky-1974}, \textit{schemata} \cite{rumelhart-1975}, and \textit{scripts} \cite{schank-abelson-1977}. 
In the linguistics and computational linguistics communities, frame semantics \cite{fillmore-1982} uses frames as the central representation of word meaning, culminating in the development of FrameNet \cite{baker-etal-1998}, which contains over 1000 manually annotated frames. A similarly rich lexical resource is the MindNet project \cite{richardson-etal-1998}.
Our notion of frame is related to these representations, but there are also subtle differences.
For example, Minsky's frame emphasizes {\em inheritance}, which we do not model in this paper. (It should be a straightforward extension: using the split-and-merge approach, ProFinder already produces a hierarchy of events and slots in learning, although currently, it simply discards the intermediate levels.) 
As in semantic role labeling, FrameNet focuses on semantic roles and does not model event or frame transitions, so the scope of its frames is often no more than an event in our model.
Perhaps the most similar to our frame is Roger Schank's scripts, which capture prototypical events and participants in a scenario such as restaurant dining. In their approach, however, scripts are manually defined, making it hard to generalize.
In this regard, our work may be viewed as an attempt to revive a long tradition in AI and linguistics, by leveraging the recent advances in computational power, NLP, and machine learning.

%% file: model.tex
\section{Probabilistic Frame Induction}
In this section, we present \texttt{ProFinder}, a probabilistic model for frame induction. 
Let $\cal F$ be a set of frames, where each frame $F=(E_F,S_F)$ comprises a unique set of events $E_F$ and slots $S_F$. 
Given a document $D$ and a word $w$ in $D$, $Z_w=(f, e)$ represents an assignment of $w$ to frame $f\in\cal F$ and frame element $e\in E_{f}\cup S_f$.
At the heart of \texttt{ProFinder} is a generative model $P_{\theta}(D, Z)$ that defines a joint distribution over document $D$ and the frame assignment to its words $Z$. 
Given a set of documents $\cal D$, frame induction in ProFinder amounts to determining the number of frames, events and slots, as well as learning the parameters $\theta$ by summing out the latent assignments $Z$ to maximize the likelihood of the document set
\[ \prod_{D \in {\cal D}} P_{\theta}(D). \]
The induced frames identify the key event structures in the document set. Additionally, \texttt{ProFinder} can also conduct event and entity extraction by computing the most probable frame assignment $Z$.
In the remainder of the section, we first present the base model for \texttt{ProFinder}. We then introduce several linguistically motivated refinements, and efficient algorithms for learning and inference in \texttt{ProFinder}.

\subsection{Base Model}
The probabilistic formulation of \texttt{ProFinder} makes it extremely flexible for incorporating linguistic intuition and prior knowledge. In this paper, we design our \texttt{ProFinder} model to capture three types of dependencies.

\paragraph{Frame transitions between clauses} A sentence contains one or more clauses, each of which is a minimal unit expressing a proposition. A clause is unlikely to straddle across different frames, so we stipulate that the words in a clause be assigned to the same frame. On the other hand, frame transitions can happen between clauses, and we adopt the common Markov assumption that the frame of a clause only depends on the clause immediately to its left. Here, sentences are ordered sequentially as they appear in the documents. Clauses are automatically extracted from the dependency parse and further decomposed into an \textit{event head} and its syntactic arguments; see the experiment section for details. 

\paragraph{Event transitions within a frame} Events tend to transition into related events in the same frame, as determined by their causal or temporal relations. Each clause is assigned an event compatible with its frame assignment (i.e., the event is in the given frame). As for frame transitions, we assume that the event assignment of a clause depends only on the event of the previous clause.

\paragraph{Emission of event heads and slot words} Similar to topics in topic models, each event determines a multinomial from which the event head is generated. E.g., a detonation event might use verbs such as {\em detonate, set off} or nouns such as {\em denotation, bombing} as its event head.
Additionally, as in USP \cite{poon-domingos-2009}, an event also contains a multinomial of slots for each of its argument types\footnote{USP generates the argument types along with events from clustering. For simplicity, in \texttt{ProFinder} we simply classify a syntactic argument into subject, object, and prepositional object, according to its Stanford dependency to the event head.}. E.g., the agent argument of a detonation event is generally the {\tt PERPETRATOR} slot of the {\tt BOMBING} frame.
Finally, each slot has its own multinomials for generating the argument head and dependency label, regardless of the event.

Formally, let $D$ be a document and $C_1,\cdots,C_l$ be its clauses, the ProFinder model is defined by
\begin{align*}
P_{\theta}(D, Z) &= P_{\tt F-INIT} (F_1) \times \prod_i P_{\tt F-TRAN} (F_{i+1}|F_i)\\
& \times P_{\tt E-INIT} (E_1|F_1) \\
& \times \prod_i P_{\tt E-TRAN} (E_{i+1}|E_i, F_{i+1}, F_i) \\
& \times \prod_i P_{\tt E-HEAD} (e_i|E_i) \\
& \times \prod_{i,j} P_{\tt SLOT} (S_{i,j} | E_{i,j}, A_{i,j}) \\
& \times \prod_{i,j} P_{\tt A-HEAD} (a_{i,j} | S_{i,j}) \\
& \times \prod_{i,j} P_{\tt A-DEP} (dep_{i,j} | S_{i,j}) 
\end{align*}
Here, $F_i, E_i$ denote the frame and event assignment to clause $C_i$, respectively, and $e_i$ denotes the event head. For the $j$-th argument of clause $i$, $S_{i,j}$ denotes the slot assignment, $A_{i,j}$ the argument type, $a_{i,j}$ the head word, and $dep_{i,j}$ the dependency from the event head. 
$P_{\tt E-TRAN} (E_{i+1}|E_i, F_{i+1}, F_i)=P_{\tt E-INIT} (E_{i+1}|F_{i+1})$ if $F_{i+1}\ne F_i$.   

Essentially, \texttt{ProFinder} combines a frame HMM with an event HMM, where the first models frame transition and emits events, and the second models event transition within a frame and emits argument slots.

\subsection{Model refinements}
The base model captures the main dependencies in event narrative, but it can be easily extended to leverage additional linguistic intuition. \texttt{ProFinder} incorporates three such refinements.

\paragraph{Background frame} Event narratives often contain interjections of general content common to all frames. For example, in newswire articles, \textsc{Attribution} is commonplace to describe who said or reported a particular quote or fact. To avoid contaminating frames with generic content, we introduce a background frame with its own events, slots, and emission distributions, and a binary switch variable $B_i \in \{BKG, CNT\}$ that determines whether clause $i$ is generated from the actual content frame $F_i$ ($CNT$) or background ($BKG$). We also stipulate that if background is chosen, the nominal frame stays the same as the previous clause.


\paragraph{Stickiness in frame and event transitions} Prior work has demonstrated that promoting topic coherence in natural-language discourse helps discourse modeling \cite{barzilay-lee-2004}. We extend \texttt{ProFinder} to leverage this intuition by incorporating a ``stickiness'' prior \cite{haghighi-vanderwende-2009} to encourage neighboring clauses to stay in the same frame. Specifically, along with introducing the background frame, the frame transition component now becomes
\begin{align}
\label{eq:f-tran}
& P_{\tt F-TRAN}(F_{i+1}|F_i,B_{i+1}) = \\
& \begin{cases}
    \mathbf{1}(F_{i+1} = F_i), & \mbox{if } B_{i+1} = BKG \\
    \begin{matrix*}[l] \beta \mathbf{1}(F_{i+1} = F_i) + \\ (1 - \beta) P_{\tt F-TRAN}(F_{i+1} | F_i), \end{matrix*}& \mbox{if }   B_{i+1} = CNT
  \end{cases} \nonumber
\end{align}
where $\beta$ is the stickiness parameter, and the event transition component correspondingly becomes
\begin{align}
\label{eq:e-tran}
& P_{\tt E-TRAN} (E_{i+1}|E_i,F_{i+1},F_i,B_{i+1}) = \\
& \begin{cases}
    \mathbf{1}(E_{i+1} = E_i), & \mbox{if } B_{i+1} = BKG \\
    P_{\tt E-TRAN} (E_{i+1}|E_i), & \begin{matrix*}[l]\mbox{if } B_{i+1} = CNT, F_i = F_{i+1} \end{matrix*}\\
    P_{\tt E-INIT}(E_{i+1}), & \begin{matrix*}[l]\mbox{if } B_{i+1} = CNT, F_i \neq F_{i+1} \end{matrix*} \\
  \end{cases} \nonumber
\end{align}

\paragraph{Argument dependencies as caseframes} As noticed in previous work such as \newcite{chambers-jurafsky-2011}, the combination of an event head and a dependency relation often gives a strong signal of the slot that is indicated. For example, $bomb>nsubj$ often indicates a \texttt{PERPETRATOR}. Thus, rather than simply emitting the dependency from the event head to an event argument $dep_{i,j}$, our model instead emits the pair of event head and dependency relation, which we call a caseframe following \newcite{bean-riloff-2004}.

\begin{figure}
	\centering
		\includegraphics[scale=0.45]{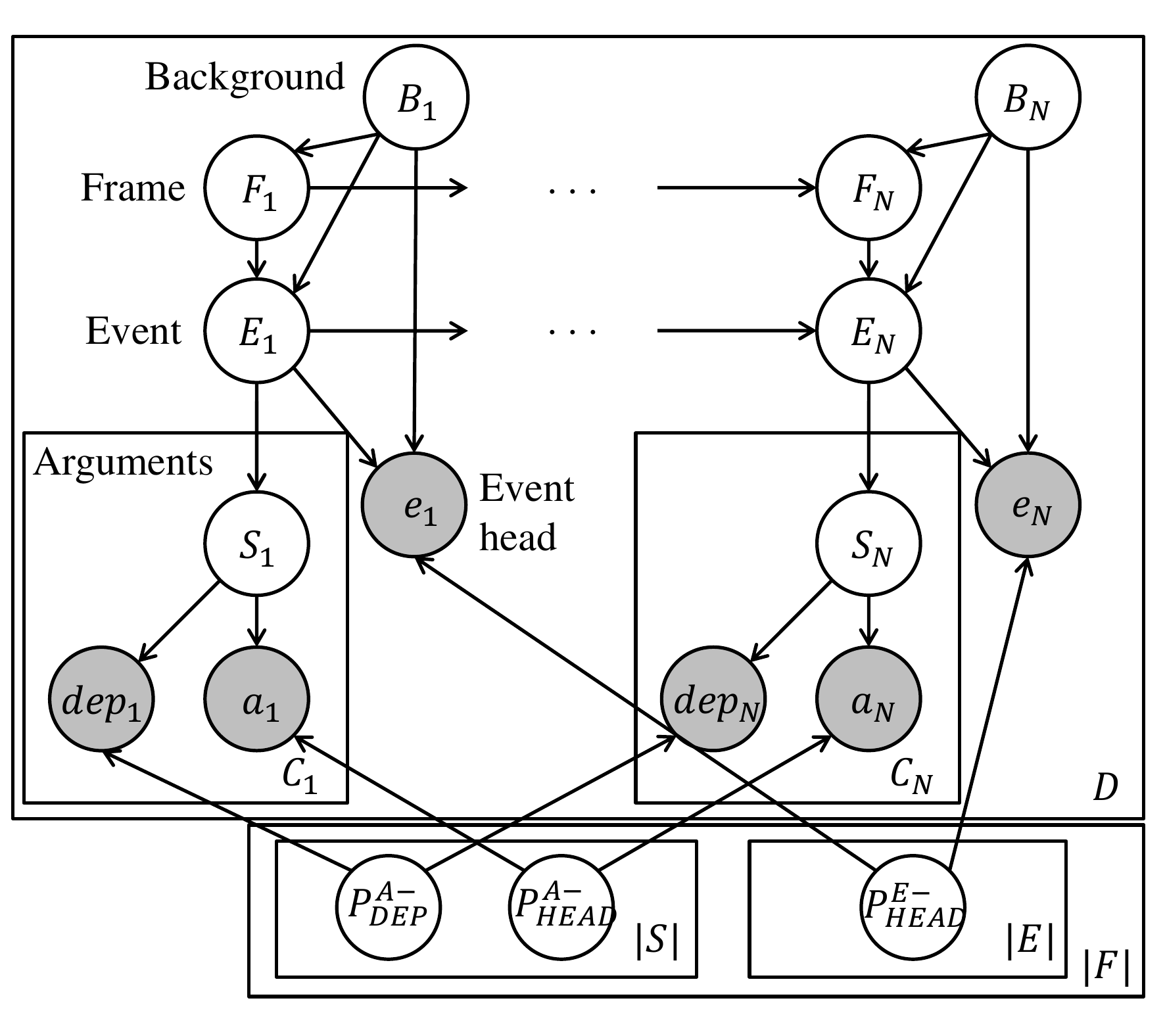}
	\caption{Graphical representation of our model. Hyperparameters, the stickiness factor, and the frame and event initial and transition distributions are not shown for clarity.}
	\label{fig:model}
\end{figure}

\subsection{Full generative story}
To summarize, the distributions that are learned by our model are the default distributions $P_{\tt BKG}(B)$, $P_{\tt F-INIT}(F)$, $P_{\tt E-INIT}(E)$, the transition distributions $P_{\tt F-TRAN}(F_{i+1}|F_i)$, $P_{\tt E-TRAN}(E_{i+1}|E_i)$, and the emission distributions $P_{\tt SLOT} (S|E,A,B)$, $P_{\tt E-HEAD}(e|E,B)$, $P_{\tt A-HEAD}(a|S)$, $P_{\tt A-DEP}(dep|S)$. We used additive smoothing with uniform Dirichlet priors for all the multinomials. The overall generative story of our model is as follows:

\begin{compactenum}
 \item Draw a Bernoulli distribution for $P_{\tt BKG}(B)$
 \item Draw the frame, event, and slot distributions
 \item Draw an event head emission distribution $P_{\tt E-HEAD}(e|E,B)$ for each frame including the background frame
 \item Draw event argument lemma and caseframe emission distributions for each slot in each frame including the background frame
 \item For each clause in each document, generate the clause-internal structure.
\end{compactenum}

The clause-internal structure at clause $i$ is generated by the following steps:
\begin{compactenum}
   \item Generate whether this clause is background ($B_i \in \{CNT, BKG\} \sim P_{\tt BKG}(B)$)
   \item Generate the frame $F_i$ and event $E_i$ from $P_{\tt F-INIT}(F)$, $P_{\tt E-INIT}(E)$, or according to equations~\ref{eq:f-tran} and \ref{eq:e-tran}
   \item Generate the observed event head $e_i$ from $P_{\tt E-HEAD} (e_i|E_i)$.
   \item For each event argument:
   \begin{compactenum}
    \item Generate the slot $S_{i,j}$ from $P_{\tt SLOT} (S|E,A,B)$.
    \item Generate the dependency/caseframe emission $dep_{i,j} \sim P_{\tt A-DEP}(dep|S)$ and the lemma of the head word of the event argument $a_{i,j} \sim P_{\tt A-HEAD}(a|S)$.
   \end{compactenum}
 \end{compactenum}

\subsection{Learning and Inference}
Our generative model admits efficient inference by dynamic programming. In particular, after collapsing the latent assignment of frame, event, and background into a single hidden variable for each clause, the expectation and most probable assignment can be computed using standard forward-backward and Viterbi algorithms.

Parameter learning can be done using EM by alternating the computation of expected counts and the maximization of multinomial parameters. In particular, \texttt{ProFinder} used incremental EM, which has been shown to have better and faster convergence properties than standard EM \cite{liang-klein-2009}.

Determining the optimal number of events and slots is challenging. One solution is to adopt non-parametric Bayesian methods by incorporating a hierarchical prior over the parameters (e.g., a Dirichlet process). However, this approach can impose unrealistic restrictions on the model choice and result in intractability which requires sampling or approximate inference to overcome. Additionally, EM learning can suffer from local optima due to its non-convex learning objective, especially when dealing with a large number hidden states without a good initialization. 

To address these issues, we adopt a novel application of the split-merge method previously used in syntactic parsing for inferring refined latent syntactic categories \cite{petrov-etal-2006}.
Specifically, we initialize our model such that each frame is associated with one event and two slots. Then, after a number of iterations of EM, we split each event and slot in two along with their probability, and duplicate the associated emission distributions. We then add some perturbation to break symmetry. After splitting, we merge back a proportion of the newly split events and slots that result in the least improvement in the likelihood of the training data. For more details on split-merge, see \cite{petrov-etal-2006}

By adjusting the number of split-merge cycles and the merge parameters, our model learns the number of events and slots in a dynamical fashion that is tailored to the data.
Moreover, our model starts with a small number of frame elements, which reduces the number of local optima and make initial learning easier. After each split, the subsequent learning starts with (a perturbed version of) the previously learned parameters, which makes a good initialization that is crucial for EM.
Finally, it is also compatible with the hierarchical nature of events and slots. For example, slots can first be coarsely split into persons versus locations, and later refined into subcategories such as perpetrators and victims.

%% file: experiment.tex
\section{MUC-4 Entity Extraction Experiments}

We first evaluate our model on a standard entity extraction task, using the evaluation settings from \newcite{chambers-jurafsky-2011} to enable a head-to-head comparison. Specifically, we use the MUC-4 data set \cite{muc4-1992}, which contains 1300 training and development documents on terrorism in South America, with 200 additional documents for testing. MUC-4 contains four templates: attack, kidnapping, bombing, and arson.\footnote{Two other templates have negligible counts and are ignored as in \newcite{chambers-jurafsky-2011}.} All templates share the same set of predefined slots, with the evaluation focusing on the following four: perpetrator, physical target, human target, and instrument. 

For each slot in a MUC template, the system first identified an induced slot that best maps to it by $F_1$ on the development set.
As in \newcite{chambers-jurafsky-2011}, template is ignored in final evaluation. 
So the system merged the induced slots across all templates to calculate the final scores. 
Correctness is determined by matching head words, and slots marked as optional in MUC are ignored when computing recall. All hyper-parameters are tuned on the development set\footnote{We will make the parameter settings used in all experiments publicly available.}.

\paragraph{Document classification} The MUC-4 dataset contains many documents that contain words related to MUC slots (e.g., \textit{plane} and \textit{aviation}), but are not about terrorism. 
To reduce precision errors, Chambers and Jurafsky's (2011) (henceforth, C\&J) \nocite{chambers-jurafsky-2011} first filtered irrelevant documents based on the specificity of event heads to learned frames. To estimate the specificity, they used additional data retrieved from a large external corpus.
In ProFinder, however, specificity can be easily estimated using the probability distributions learned during training. In particular, we define the probability of an event head in a frame $j$:
\begin{align}
P_F(w) = \sum_{E_F \in F} P_{\tt E-HEAD} (w|E) / |F|,
\end{align}
and the probability of a frame given an event head:
\begin{align}
P(F|w) = P_F(w) / \sum_{F' \in \cal F} P_{F'}(w).
\end{align}
We then follow the rest of \newcite{chambers-jurafsky-2011} to score each learned frame with each MUC document, mapping a document to a frame if the average $P_F(w)$ in the document is above a threshold and the document contains at least one \textit{trigger word} $w'$ with $P(F|w') > 0.2$. The threshold and the induced frame were determined on the development set, which were then used to filter irrelevant documents in the test set.

\begin{table}
	\centering
		\begin{tabular}{llll}
			\textbf{Unsupervised methods} & $P$ & $R$ & $F_1$ \\
			\texttt{ProFinder} (This work) & 32 & \textbf{37} & \textbf{34} \\
			\newcite{chambers-jurafsky-2011} & \textbf{48} & 25 & 33 \\
			 \textbf{With extra information} &  &  &  \\
			\texttt{ProFinder} +doc. classification & 41 & \textbf{44} & \textbf{43} \\
			C\&J 2011 +granularity & \textbf{44} & 36 & 40
		\end{tabular}
	\caption{Results on MUC-4 entity extraction. C\&J~2011 +granularity refers to their experiment in which they mapped one of their templates to five learned clusters rather than one.}
	\label{tab:muc4-results}
\end{table}

\paragraph{Results} Compared to C\&J, \texttt{ProFinder} is conceptually much simpler, involving a single probabilistic model, with standard learning and inference algorithms. In particular, it did not require multiple processing steps or customized similarity metrics; rather, it only used the data within MUC-4. In contrast, C\&J required additional text to be retrieved from a large external corpus (Gigaword \cite{graff-etal-2005}) for each event cluster,
yet \texttt{ProFinder} nevertheless was able to outperform C\&J on entity extraction, as shown in Table~\ref{tab:muc4-results}.
Our system achieved good recall but was hurt by the lower precision. 
We investigated the importance of document classification by only extracting from the gold-standard relevant documents (+doc. classification), which led to a substantial improvement in precision, suggesting possible further improvement by better document classification.
Also unlike C\&J, our system does not currently make use of coreference information.

\begin{figure}
  \centering
  \begin{tabular}{p{0.22\textwidth}|p{0.22\textwidth}}
  \multicolumn{2}{c}{\textbf{Frame: Terrorism}} \\
	\multicolumn{1}{c|}{\textbf{Event: Attack}} & \multicolumn{1}{c}{\textbf{Event: Discussion}}\\
	\textit{report, participate, kidnap, kill, release} & 
	\textit{hold, meeting, talk, discuss, investigate} \\ \hline
	
	\multicolumn{1}{c|}{\textbf{Slot: Perpetrator}} & \multicolumn{1}{c}{\textbf{Slot: Victim}}\\
	\multicolumn{1}{c|}{\textsc{Person/Org}} & \multicolumn{1}{c}{\textsc{Person/Org}}\\
\textbf{Words:} \textit{guerrilla, police, source, person, group} & 
\textbf{Words:}	\textit{people, priest, leader, member, judge} \\
\textbf{Caseframes:} \textit{report$>$nsubj, kidnap$>$nsubj, kill$>$nsubj, participate$>$nsubj, release$>$nsubj} & 
\textbf{Caseframes:} \textit{kill$>$dobj, murder$>$dobj, release$>$dobj, report$>$dobj, kidnap$>$dobj} \\	
  	\end{tabular}
	\caption{A partial frame learned by ProFinder from the MUC-4 data set, with the most probable emissions for each event and slot. Labels are assigned by the authors for readability.}
	\label{fig:muc-learned}
\end{figure}

Figure~\ref{fig:muc-learned} shows part of a frame that is learned by \texttt{ProFinder}, including some of the standard MUC slots and events. Our method also finds events not annotated in MUC, such as the discussion event. Other interesting events and slots that we noticed include an arrest event (\textit{call, arrest, express, meet, charge}), a peace agreement slot (\textit{agreement, rights, law, proposal}), and an authorities slot (\textit{police, government, force, command}). The background frame was able to capture many verbs related to reporting, such as \textit{say, continue, add, believe}, although it missed \textit{report}.

%% file: experiment2.tex
\section{Evaluating Frame Induction Using Guided Summarization Templates}

One issue with the MUC-4 evaluation is the limited variety of templates and entities that are available. Moreover, this data set was specifically developed for information extraction and questions remain whether our approach can generalize beyond it. We thus conducted a novel evaluation using the TAC guided summarization data set, which contains a wide variety of frames and topics. Our evaluation corresponds to a view of summarization as extracting structured information from the source text, and highlights the connection between summarization and information extraction \cite{white-etal-2001}.

\begin{figure}
	\centering
	\begin{tabular}{rp{0.4\textwidth}}
	(a) &
\textbf{Accidents and Natural Disasters:}\\
&WHAT: what happened\\
&WHEN: date, time, other temporal markers\\
&WHERE: physical location\\
&WHY: reasons for accident/disaster\\
&WHO\_AFFECTED: casualties...\\
&DAMAGES: ... caused by the disaster\\
&COUNTERMEASURES: rescue efforts...\\

(b)
& \textit{(\textsc{When} During the night of July 17,) (\textsc{What} a 23-foot $<$\textsc{What} tsunami) hit the north coast of Papua New Guinea (PNG)$>$, (\textsc{Why} triggered by a 7.0 undersea earthquake in the area).}
\\
(c) & 
\textsc{When}: \textit{night} \quad\quad \textsc{What}: \textit{tsunami, coast} \\ & \textsc{Why}: \textit{earthquake} \\
\end{tabular}

	\caption{An example of (a) a frame from the TAC Guided Summarization task with abbreviated slot descriptions, (b) an annotated TAC contributor, and (c) the entities that are extracted for evaluation.}
	\label{fig:tac-example}
\end{figure}

\paragraph{Data preparation} We use the TAC 2010 guided summarization data set for our experiments \cite{owczarzak-dang-2010}. This data set provides templates as defined by the task organizers and contains 46 document clusters in five domains, with each cluster comprising 20 documents on a specific topic. Eight human-written model summaries are provided for each document cluster. As part of the Pyramid evaluation method \cite{nenkova-passonneau-2004}, these summaries have been manually segmented and labeled with slots from the corresponding template for each segment (Figure~\ref{fig:tac-example})\footnote{The full set of slots is available at \url{http://www.nist.gov/tac/2010/Summarization/Guided-Summ.2010.guidelines.html}}.

We first considered defining the task as extracting entities from the source text, but this annotation is not available in TAC, and pilot studies suggested that it required nontrivial effort to train average users to conduct high-quality annotation reliably. 
We thus defined our task as extracting entities from the model summaries instead. As mentioned earlier, TAC slot annotation is available for summaries. Furthermore, using the summary text has the advantage that slots that are considered important in the domain naturally appear more frequently, whereas unimportant text is filtered out.

Each span that is labeled by a slot is called a \textit{contributor}.
We convert the contributors into a form that is more like the previous MUC evaluation, so that we can fairly compare against previous work like C\&J that were designed to extract information into that form. Specifically, we extract the head lemma from all the maximal noun phrases found in the contributor. Like in MUC-4, we count a system-extracted noun phrase as a match if this head word matches and is extracted from the same document (i.e., summary). This process can lead to noise, as the meaning of some contributors depend on a larger phrasal unit than a noun phrase, but this heuristic normalizes the representations of the contributors so that they are amenable to our evaluation. We leave the denoising of this process to future work, and believe it should be feasible by crowdsourcing.

\paragraph{Method and experiments}
The induced entity clusters are mapped to the TAC slots in the TAC frames according to the best $F_1$ achieved for each TAC slot. However, one issue is that many TAC slots are more general than the type of slots found in MUC. For example, slots like \textsc{Why} and \textsc{Countermeasures} likely correspond to multiple slots at the granularity of MUC. Thus, we map the $N$-best induced slots to TAC slots rather than the 1-best, for $N$ up to 5.
We train \texttt{ProFinder} and a reimplementation of C\&J on the 920 full source texts of TAC 2010, and test them on the 368 model summaries. We do not provide C\&J's model with access to external data, in order to create fair comparison conditions to our model. We also eliminate a sentence relevance classification step from C\&J, and the document relevance classification step from both models, because all sentences in the summary text are expected to be relevant. We tune C\&J's clustering thresholds and the parameters to our model by two-fold cross validation on the summaries, and assume gold summary classification into the five topic categories defined by TAC.

\begin{table}
	\centering
		\begin{tabular}{llll|lll}
		& \multicolumn{3}{c}{\textit{1-to-1}} & \multicolumn{3}{c}{\textit{5-to-1}} \\
			\textbf{Systems} & $P$ & $R$ & $F_1$ & $P$ & $R$ & $F_1$\\
			\texttt{ProFinder} & 24 & \textbf{25} & \textbf{24} & 21 & \textbf{38} & \textbf{27} \\
			C\&J & \textbf{58} & 6.1 & 11 & \textbf{50} & 12 & 20 \\
		\end{tabular}
	\caption{Results on TAC 2010 entity extraction with $N$-to-1 mapping for $N=1$ and $N=5$. Intermediate values of $N$ produce intermediate results, and are not shown for brevity. }
	\label{tab:tac2010-results}
\end{table}

\paragraph{Results} The results on TAC are shown in Table~\ref{tab:tac2010-results}. The overall results are poorer than for the MUC-4 task, but this task is harder given the greater diversity in frames and slots to be induced. Like in the previous evaluation, our system is able to outperform C\&J in terms of recall and $F_1$, but not precision. C\&J's method produces many small clusters, which makes it easy to achieve high precision. The $N$-to-1 mapping procedure can also be seen to favor their method over ours, many small clusters with high precision can be selected to greatly improve recall, which is indeed the case. However, \texttt{ProFinder} with 1-to-1 mapping outperforms C\&J even with 5-to-1 mapping.

%% file: summary.tex
\section{Conclusion}
We have presented the first probabilistic approach to frame induction and shown that it achieves state-of-the-art results on end-to-end entity extraction in standard MUC and TAC data sets. Our model is inspired by recent advances in unsupervised semantic induction and in content modeling in summarization, and is easy to extend. We would like to further investigate frame induction evaluation, for example to evaluate event clustering in addition to the slots and entities. 

\eat{
We have proposed the first generative probabilistic model for frame induction, and shown that it achieves state-of-the-art results on entity extraction for two data sets. Our model, inspired by content models in summarization, demonstrates the close link between IE and NLG, especially automatic summarization.

We see a number of promising future directions of research. First, to improve the model itself, we could incorporate more sources of knowledge such as coreference information, external unlabeled text, or better modeling of the discourse structure, which our probabilistic formulation makes easy to do. While we have made a case for unsupervised methods and the importance of robustness across domains, our method is also amenable to semi-supervised or supervised learning if the particular domain is known and contains labeled data. Finally, we would like to further investigate frame induction evaluation, for example to evaluate event clustering in addition to the slots and entities.
}